# Learning Unsplit-field-based PML for the FDTD Method by Deep Differentiable Forest

Yingshi Chen, Naixing Feng

*Abstract*—Alternative unsplit-filed-based absorbing boundary condition (ABC) computation approach for the finite-difference time-domain (FDTD) is efficiently proposed based on the deep differentiable forest. The deep differentiable forest (DDF) model is introduced to replace the conventional perfectly matched layer (PML) ABC during the computation process of FDTD. The field component data on the interface of traditional PML are adopted to train DDF-based PML model. DDF has the advantages of both trees and neural networks. Its tree structure is easy to use and explain for the numerical PML data. It has full differentiability like neural networks. DDF could be trained by the powerful techniques from deep learning. So compared to the traditional PML implementation, the proposed method can greatly reduce the size of FDTD physical domain and the calculation complexity of FDTD due to the novel model which only involves the one-cell thickness of boundary layer. Numerical simulations have been carried out to benchmark the performance of the proposed approach. Numerical results illustrate that the proposed method can not only easily replace the traditional PML, but also be integrated into the FDTD computation process with satisfactory numerical accuracy and compatibility to the FDTD.

## I. INTRODUCTION

The stretched coordinate perfectly matched layer (SC-PML) [1], presented firstly by Chew and Weedon, is a highly effective absorption boundary condition (ABC) material to truncate the finite-difference time-domain (FDTD) computational domains and has the advantage of simple implementation in the edges and corners of PML domains. However, the implementation of the presented SC-PML formulations in [1] needed splitting the field components. Next, based on the digital signal processing (DSP) technique and the auxiliary differential equation (ADE) method, two simple and efficient PML formulations, respectively, were proposed [2, 3] instead of the splitting of the field components. Recently, based on the ADE and the DSP techniques, another two SC-PML algorithms [4, 5] using the memory-minimized method (Tri-M) were presented and their main advantages are that only one auxiliary variable is needed in some edges and all corners of the PML domains. Even if memory requirements, proposed by Li [4, 5], are saved significantly as compared with those in published papers [2, 3], they can only produce results as accurate as those in [2, 3].

To the best of our knowledge, however, the various SC-PML algorithms are not effective at attenuating the evanescent waves in [2-5]. In order to conquer this limit, the complex frequency shifted PML (CFS-PML) [6] has been presented and attracted overwhelming attention due to the fact that the CFS-PML is efficient in terms of absorbing the low-frequency evanescent waves and decreasing late-time reflections. In [7, 8], modified SC-PMLs, based on ADE and DSP methods, respectively, were proposed detailedly to effectively implement the CFS-PML.

While the CFS-PMLs are effective at absorbing evanescent waves, they are very inefficient in attenuating low-frequency propagating waves, which has been demonstrated and analyzed by J. Jin [9-10]. Recently, following Correia's seminal work, effective implementations of the multi-pole PMLs based on the unsplit-field PML formulations [11-14] have been proposed for truncating the FDTD domains. However, to reduce reflection and reach the best possible performance of PML, the thickness of PML always has to be increased so that the residue error can be reduced.

To our knowledge, machine-learning-based methods [15] have been commonly applied in the engineering and science with the booming development of computer science. Moreover, we have found its successful applications [16-19]. As well known, the machine learning extract the potential mapping disciplinarian from training data with the same pattern, and then predict a new output [15]. There have been works about using the artificial neural network (ANN) model to directly solve the FDTD [20]. However, the unavoidably accumulated error during the ANN calculation process makes directly solving FDTD by ANN an unwise choice. Considering the computing process within PML domains, it repeatedly achieves local fields based on local and neighboring fields in the current and previous step. Therefore, the machine learning method can be adopted and embedded into FDTD to replace the computation process of the PML.

In this paper, we propose a novel deep differentiable forest based ABC model (DDF-ABC) to replace the traditional PML. Differentiable decision tree [21-23] is a novel structure with the advantages of both trees and neural networks. Deep neural networks are the best machine learning method now. However tree model still has some advantages. Its structure is very simple, easy to use and explain the decision process. Especially for numerical data in the calculation of PML, tree-based models usually have better accuracy than deep networks, which is

Manuscript received

This work is supported in part by the National Nature Science Foundation of China under Grant 61901274, by the Shenzhen Science and Technology Innovation Committee under Grant JCYJ20190808141818890, and by the Guangdong Natural Science Foundation under Grant 2020A1515010467.

Naixing Feng is with Institute of Microscale Optoelectronics, Shenzhen University, Shenzhen 518060, China. (E-mail: fengnaixing@gmail.com).

Yingshi Chen is with the Institute of Electromagnetics and Acoustics, and Department of Electronic Science, Xiamen University, Xiamen 361005, China.

verified by many kaggle competitions and real applications. While keeping simple tree structure, the differentiable trees also have full differentiability like neural networks. Therefore we could train it with many powerful optimization algorithms (SGD, Adam, et al., just like the training of deep CNN. It could use the end-to-end learning mode. No need of many works to preprocess data.

## II. DEEP DIFFERENTIABLE FOREST

To describe the problem more concisely, we give the following formulas and symbols: For a dataset with N samples $X = \{x\}$ and its target $Y = \{y\}$. Each $x$ has M attributes, $x = [x_1, x_1, \cdots, x_M]^T$. The differentiable forest model would learn an ensemble of differentiable decision trees $\{T^1, T^2, T^3, \ldots, T^K\}$ to minimize the loss between the target $y$ and prediction $\hat{y}$.

$$\hat{y} = \frac{1}{K}\sum_{h=1}^{K} T^h(x) \quad (1)$$

As figure 1 shows, each node in the tree has one gating function with some parameters. It has two child nodes, which are represented as $\{\text{left}, \text{right}\}$ or $\{\swarrow, \searrow\}$. The root node would redirect input $x$ to both $\{\text{left}, \text{right}\}$ with probabilities calculated by the gating function $g$. Formula (2) gives the general definition of gating function, where $A \in R^M$ is a learnable weight parameter for each attribute of $x$, $b$ is a learnable threshold. $\sigma$ would map $Ax - b$ to probability between [0,1], for example, the sigmoid function.

$$g(A, x, b) = \sigma(Ax - b) \quad (2)$$

The sample $x$ would be directed to each nodal j with probability $p_j$. And finally, the input $x$ would reach all leaves. For a tree with depth $d$, we represent the path as $\{n_1, n_2, \cdots, n_d\}$, where $n_1$ is the root node and $n_d$ is the leaf node j. $p_j$ is just the product of the probabilities of all nodes in this path:

$$p_j = \prod_n g_n \text{ where } n \in \{n_1, n_2, \cdots, n_d\} \quad (3)$$

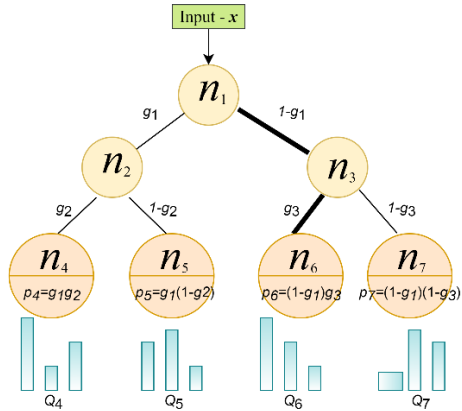

Figure 1 differentiable tree and its probability and response at leaf nodes. In this sample, The input $x$ would reach $n_3$ with probability 1-$g_1$ and reach $n_6$ with probability (1-$g_1$) $g_3$.

Let the response at leaf node j represented by learnable parameter $Q_j$. Then the output of the tree is just the probability average of these responses.

$$\hat{y}(x) = \sum_j p_j Q_j(x) \text{ where } j \text{ is the leaf} \quad (4)$$

A single tree is a very weak learner, so we should merge many trees to get higher accuracy, just like the random forest or other ensemble learning method. Let $\Theta$ represents all parameters $(A, b, Q)$, then the final loss would be represented by the general form of (5)

$$L(\Theta: x, y) = \frac{1}{K}\sum_{h=1}^{K} L^h(\Theta: x, y) = \frac{1}{K}\sum_{h=1}^{K} L^h(A, b, Q: x, y) \quad (5)$$

In the case of classification problem, the classical function of $L$ is cross-entropy. For regression problem, $L$ maybe mse, mae, huber loss or others. To minimize the loss (5), we use stochastic gradient descent (SGD) method to train this model.

## III. LEARNING THE PML

Based on the general loss function defined in (5), we use stochastic gradient descent method [26, 27] to reduce the loss. As formula (6) shows, update all parameters $\Theta$ batch by batch:

$$\Theta^{(t+1)} = \Theta^t - \eta\frac{\partial L}{\partial \Theta}(\Theta^t; \mathcal{B}) = \Theta^t - \frac{\eta}{|\mathcal{B}|}\sum_{(x,y)\in\mathcal{B}}\frac{\partial L}{\partial \Theta}(\Theta^t; x, y) \quad (6)$$

where $\mathcal{B}$ is the current batch, $\eta$ is the learning rate, $(x, y)$ is the sample in current batch.

This is similar to the training process of deep learning. Some hyperparameters (batch size, learning rate, weight decay, drop out ratio…) need to be set. All the training skills and experience from deep learning could be used. For example, the batch normalization technique, drop out layer, dense net…We find QHAdam [26] would get a few higher accuracy than Adam [27]. Consequently, QHAdam is the default optimization algorithm in QuantumForest.

**Input**
  training, validing and testing dataset
**Initialization**
  Init the sparse attention $A$
  Init response $Q$ at each leaf nodes
  Init threshhold values $b$
**repeat**
  For each batch $\mathcal{B} \subseteq X$:
    Forward and get loss
    $$L(\Theta: x, y) = \frac{1}{K}\sum_{h=1}^{K} L^h(A, b, Q: x, y)$$
    Backpropagate to get the gradient
    $(\Delta A, \Delta b, \Delta Q) = \Delta L$
    Update the parameters
    $A \leftarrow A - \eta\Delta$
    $b \leftarrow b - \eta\Delta b$
    $Q \leftarrow Q - \eta\Delta Q$
    Evalue loss at valiending dataset
**until** convergence

Algorithm 1 Learning DDF-based ABC model.

## IV. NUMERICAL RESULTS

To simplify the problem, but without loss of generality, we model a 2D TE-polarized electromagnetic wave interaction with an infinitely long perfectly electric conductor (PEC) sheet with the finite width to validate the proposed formulations. Fig. 2 shows the FDTD grid geometry used in this simulation.

The space is discretized with the FDTD lattice with $\Delta x = \Delta y = 1$ mm and time step is $\Delta t = 1.1785$ ps $\approx 0.5 \times$ CFL. The FDTD computational domains consist of a 100-cell wide PEC sheet surrounded by free space. For a conventional PML case, 10-cell thickness PML is used to terminate the physical domain and is placed only 3 cells away from the PEC sheet in all directions, while the area contains only 1-cell DDF-based PML layer at each edge for a DDF-based PML case.

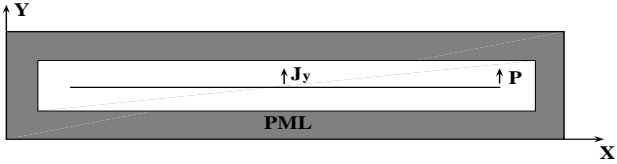

Fig. 2 shows the FDTD grid geometry in this simulation.

A y-polarized line electric current source, infinitely long in the z direction, is placed at the center and excited with a differentiated Gaussian pulse given by

$$J_y(t) = -2\frac{t-t_0}{t_w}\exp\left[-\left(\frac{t-t_0}{t_w}\right)^2\right] \quad (7)$$

where $t_w = 26.53$ ps and $t_0 = 4t_w$. The y-component of the electric field is measured at the point P, where we expect very strong evanescent waves to appear. The relative reflection error (in dB) versus time is computed at the observation point P by using

$$R_{dB}(t) = 20\log_{10}\left(\frac{|E_y^R(t) - E_y^T(t)|}{|E_{y\_\max}^R|}\right) \quad (8)$$

where $E_y^T(t)$ represents the time-dependent discrete electric field of the observation point , $E_y^R(t)$ is a reference solution based on a larger grid, and $E_{y\_\max}^R$ represents the maximum value of the reference solution over the full time simulation. The reference grid is sufficiently large such that there are no reflections from its outer boundaries during 1500 time steps which are well past the steady-state response.

In this example, to train our DDF-ABC model, we are going to use thousands of groups of data as the dataset.

## V. CONCLUSION

In this work, an efficient machine learning method is introduced to construct novel and unsplit-field ABC for the FDTD implementation. We here propose the DDF -based PML to replace the conventional PML. Compared to the traditional PML ABC, the proposed model merely involves the boundary layer with one cell to attenuate the propagating wave and greatly reduce the size of computational region. With satisfactory accuracy and low-computation complexity, the proposed new model can be successfully replace the traditional PML ABC, and exhibit flexible integrity for the FDTD solving process.